\documentclass[conference]{IEEEtran}
\IEEEoverridecommandlockouts
\usepackage{cite}
\usepackage{amsmath,amssymb,amsfonts}
\usepackage{algorithmic}
\usepackage{graphicx}
\usepackage{textcomp}
\usepackage{xcolor}
\usepackage[pagebackref=true,breaklinks=true,colorlinks,bookmarks=false]{hyperref}

\def\BibTeX{{\rm B\kern-.05em{\sc i\kern-.025em b}\kern-.08em
    T\kern-.1667em\lower.7ex\hbox{E}\kern-.125emX}}
\begin{document}

\title{FIS-OT: Feature-Induced Optimal Transport for Unsupervised Action Segmentation\\
\thanks{$^{*}$Equal contribution}%
\thanks{$^{\ddag}$Corresponding authors}
}
\author{%
  \IEEEauthorblockN{%
    Linxiang Peng\textsuperscript{1,2}$^{*}$,
    Xinyao Qin\textsuperscript{1,2}$^{*}$,
    Jinhan Li\textsuperscript{1,2},
    Di Yang\textsuperscript{1,2,3}$^{\ddag}$,
    Jiangtao Wang\textsuperscript{1,2,3,4}$^{\ddag}$
    \vspace{0.cm}
  }
  \IEEEauthorblockA{%
   \textsuperscript{1}Suzhou Institute for Advanced Research, University of Science and Technology of China\\
   \textsuperscript{2}School of Artificial Intelligence and Data Science, University of Science and Technology of China\\
    \textsuperscript{3}Suzhou Big Data \& AI Research and Engineering Center\\
    \textsuperscript{4}Shanghai Key Laboratory of Data Science
  }
  \small{\{plx, qxy15196216301, ljh340104\}@mail.ustc.edu.cn} \\ 
  \small{\{di.yang, wangjiangtao\}@ustc.edu.cn}
}

\maketitle
\vspace{-5pt}
\begin{abstract}
Unsupervised action segmentation is a challenging task. It involves finding action categories and boundaries in videos without labels. Existing Optimal Transport (OT) methods use global constraints. This causes them to overlook the use of local information.  Furthermore, existing Optimal transport architectures are prone to confirmation bias because they overly trust the pseudo-labels they generate. This causes models to learn from noise in the early training stages. To address these issues, we propose FIS-OT. It is a novel Feature-Induced Structured Optimal Transport framework. First, we introduce a Feature Enhanced Generator (FEG) module. It serves as an internal regularizer. By using triplet loss, FEG captures local consistency. It provides robust supervision that is independent of noisy pseudo-labels. Second, we propose a Feature-Induced Residual Structural Prior. This combines a fixed temporal backbone with dynamic feature similarities. This design ensures temporal continuity. It also allows the solver to adapt to complex action structures. Finally, we establish a cyclic optimization loop. This aligns local feature learning with global structural alignment. Extensive experiments on the three datasets show the effectiveness of our method. Code is availble at \href{https://github.com/flying05/FIS-OT}{https://github.com/flying05/FIS-OT}.
\end{abstract}

\section{Introduction}
\label{sec:intro}
In recent years, unsupervised long-form video action segmentation has attracted increasing attention\cite{1cad,2tot,3otas,4hvq,5ufsa,6asot}. This task is critical for real-world applications, including instructional video analysis, video content understanding and athletic action detection \cite{7}. Classical approaches to unsupervised action segmentation typically learn frame-wise embeddings and subsequently cluster them into action segments\cite{8cte,9asal,10}. 

More recently, Optimal Transport (OT) based methods have emerged as a powerful tool.They leverage OT formulations to generate pseudo-labels for self-training\cite{2tot,5ufsa,6asot}.
\begin{figure}[htbp]
    \centering
    \includegraphics[width=\linewidth]{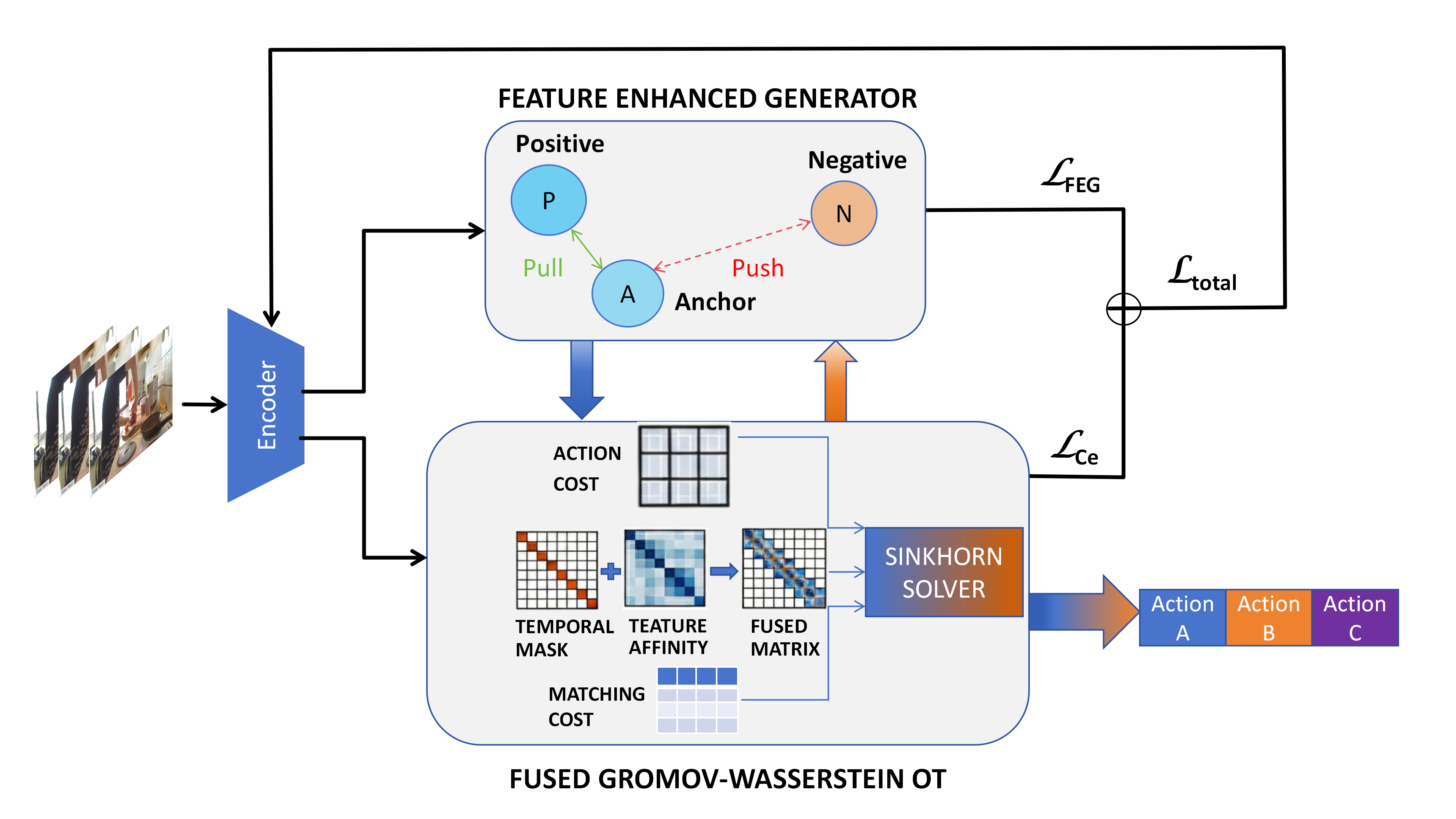}
    \caption{Illustration of the FIS-OT framework. Our model combines two key components. First, the Feature Enhanced Generator (FEG) learns local features. Second, the Fused Gromov-Wasserstein OT module handles global structure. The OT solver uses a dynamic structural prior. By mixing a temporal mask with feature affinity, it generates accurate pseudo-labels.}
    \label{fig1}
\vspace{-5pt}
\end{figure}
Among these OT-based methods, ASOT\cite{6asot}is a representitive method that is widely used. Unlike prior works, the advantage of ASOT is that it does not presuppose any temporal ordering of actions, instead, it casts segmentation as a fused Gromov–Wasserstein (FGW) problem between frames and action labels. 

However, ASOT still faces limitations in real-world videos. First, \textbf{a rigid structural prior}. The fixed temporal mask of ASOT introduces a rigid structural prior. It promotes smooth transitions but fails to reflect the true temporal dynamics inside each action. This causes short actions and rapid transitions to be smoothed out, and the model becomes biased toward long segments when frame features are noisy. Second, \textbf{over-dependency on pseudo-labels}. ASOT relies entirely on pseudo-labels from its own OT solver. When early pseudo-labels are inaccurate, the encoder updates amplify these errors. This confirmation bias persists because the model lacks an independent source of supervision. Existing methods \cite{1cad,9asal,4hvq} aim to refine features, but they do not link feature learning with the structural constraints of the OT formulation.

To address these issues, we propose FIS-OT, a novel \textbf{F}eature-\textbf{I}nduced \textbf{S}tructural \textbf{O}ptimal \textbf{T}ransport framework. FIS-OT introduces a local feature generator and a dynamic structural prior that guide each other during training. The Feature Enhanced Generator (FEG) learns temporal–semantic similarity through a KL-based triplet objective that does not depend on pseudo-labels, which improves local feature reliability and helps preserve short actions. At the global level, we propose a feature-induced residual structural prior. It combines a stable temporal backbone with a feature-driven affinity term, so the FGW solver can adjust its structure when the learned features improve. Both components use the same Gaussian kernel metric, which keeps the optimization consistent. A simple curriculum strategy further stabilizes the interaction between the local and global modules.

These designs create a cyclic optimization process that reduces confirmation bias and improves segmentation on both short and long actions. 

Our main contributions in this work are summarized as follows:
(i) We introduce a novel framework that links local feature learning with global OT-based structural alignment, addressing the rigid prior and confirmation bias issues in ASOT.
(ii) We propose a feature-induced residual structural prior that adapts temporal structure based on feature affinity, while preserving stability through a temporal backbone.
(iii) We develop a feature-enhanced generator that provides pseudo-label-independent supervision, improving fine-grained temporal–semantic consistency.
(iv) We achieve state-of-the-art performance on multiple benchmark datasets, with clear gains in boundary accuracy and short-action detection.
\section{Related Work}
\subsection{Fully- and Weakly-Supervised Action Segmentation}
Fully supervised action segmentation methods generally yield the most reliable performance but necessitate expensive frame-wise annotations. Early approaches typically modeled temporal action structures using sliding windows or Hidden Markov Models (HMM). To capture long-range dependencies, subsequent research shifted towards Temporal Convolutional Networks (TCNs), represented by MS-TCN\cite{13mstcn} and its variants, which utilize hierarchical temporal pooling. More recently, Transformer-based architectures, such as ASFormer\cite{14asformer}, have further pushed performance boundaries by leveraging self-attention mechanisms. To improve scalability and reduce annotation costs, weakly-supervised approaches have emerged. These methods rely on coarser forms of supervision, such as video-level transcripts, action sets, or timestamps. While these paradigms significantly lower the annotation burden, they still require a certain degree of prior knowledge regarding action classes or their ordering.
\vspace{-5pt}
\subsection{Unsupervised Action Segmentation}
Unsupervised methods aim to discover action patterns and semantically meaningful classes directly from long, untrimmed videos without any manual labeling. Existing frameworks can be broadly categorized into video-level and activity-level approaches based on their processing granularity.

\subsubsection{Video-level action segmentation} This category focuses on processing videos individually without predefined activity categories. Recent approaches primarily utilize clustering or boundary detection based on visual similarities. For instance, TW-FINCH\cite{15twf} incorporates temporal proximity alongside semantic similarity for hierarchical clustering, while ABD\cite{16abd} detects action boundaries by measuring adjacent frame similarities. However, these methods operate in isolation, neglecting the semantic consistency of actions across the entire dataset.

\subsubsection{Activity-level action segmentation} This domain traditionally follows a two-step pipeline: first, learning action representations in a self-supervised manner, and then clustering the learned embeddings, typically assuming a predefined number of clusters. Classical methods strongly rely on temporal regularization to model the sequential nature of activities. This paradigm has been further refined through encoder-decoder architectures, incorporating either visual reconstruction losses ( VTE\cite{17vte}) or discriminative embedding losses (UDE\cite{18ude}) to enhance clustering performance. Other approaches frame the problem as a self-supervised learning task, identifying action prototypes via auxiliary classification objectives (CAD\cite{1cad}) or distinguishing valid action orderings based on shuffled segment predictions (ASAL\cite{9asal}). Distinct from these paradigms, HVQ\cite{4hvq} reformulates the task as hierarchical vector quantization, employing multi-granularity codebooks to explicitly address the segment length bias inherent in standard clustering baselines.
\subsection{Optimal Transport for Action Segmentation}
Recently, Optimal Transport (OT) has emerged as a dominant paradigm in this domain due to its efficacy in sequence alignment and distribution matching. TOT\cite{2tot} pioneered the use of temporal OT to generate pseudo-labels for self-training, although it relies on the strict assumption of a fixed action ordering across all videos. UFSA\cite{5ufsa} relaxed this constraint but still requires prior knowledge of an estimated action order to infer the segmentation.

The current state-of-the-art method, ASOT\cite{6asot}, achieves temporally consistent segmentation without order assumptions by solving a fused Gromov-Wasserstein OT problem. Despite its effectiveness, ASOT faces two critical limitations. First, it prioritizes global structural constraints, such as rigid diagonal masks, at the expense of local information. Consequently, it overlooks fine-grained cues and struggles to detect short-duration actions. Second, it relies heavily on self-generated pseudo-labels for supervision. This leads to confirmation bias, where the model reinforces its own prediction errors during training. 

To address these challenges, we propose FIS-OT. We introduce a Feature Enhancing Generator(FEG) module to capture intrinsic local cues independent of noisy pseudo-labels. Furthermore, we employ a Feature-Induced Residual Structural mechanism to adaptively guide the global transport plan.
\section{Proposed Approach}
\begin{figure*}[t] 
    \centering
    \includegraphics[width=\textwidth]{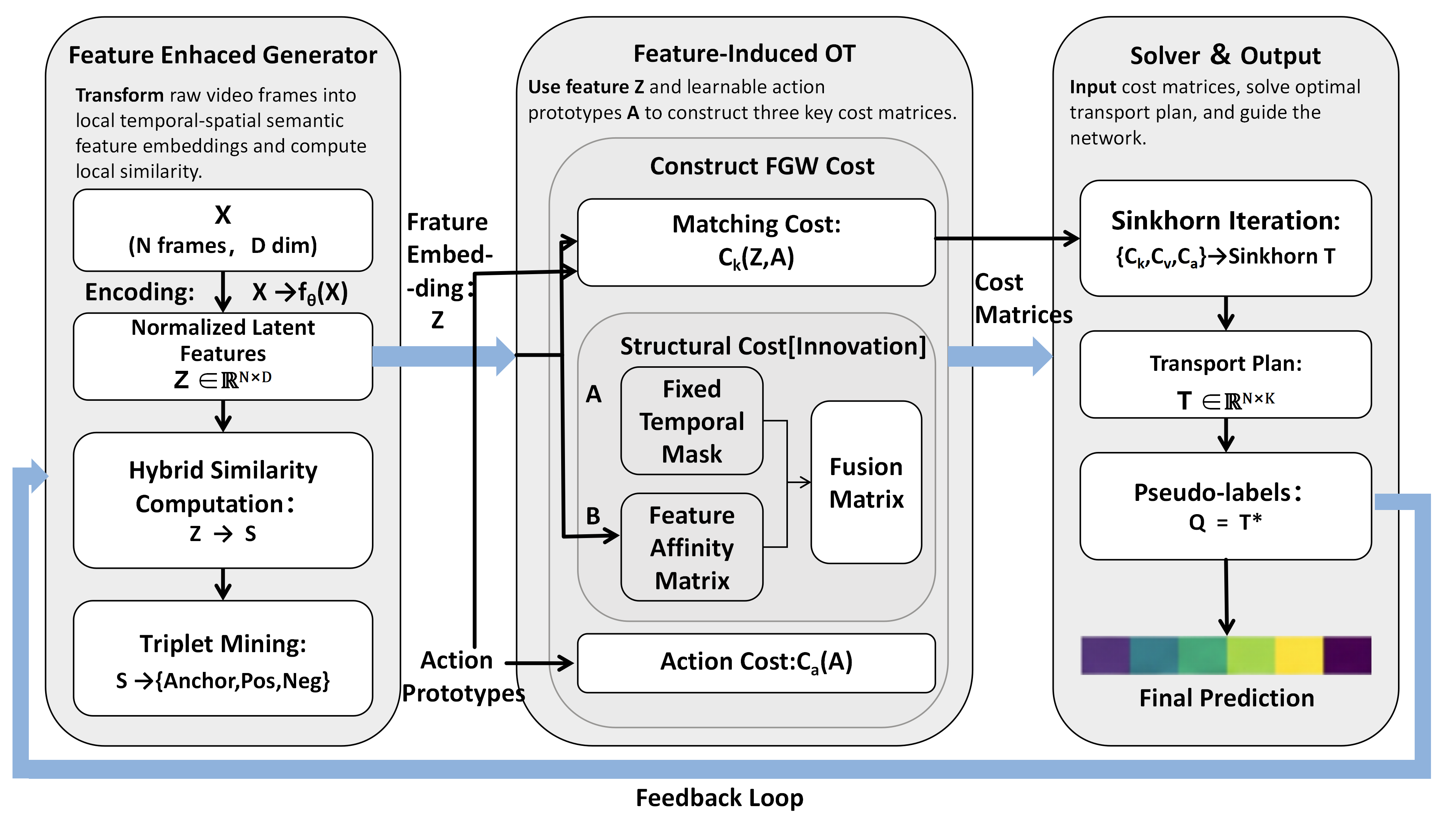}
    \caption{Overview of the FIS-OT framework. The model consists of three main parts. First, the (a) Feature Enhanced Generator extracts features from video frames and calculates local similarity. Second, the (b) Feature-Induced OT module builds the cost matrices. It combines a fixed temporal mask with a dynamic feature affinity matrix. Finally, the (c) Solver produces the optimal transport plan and pseudo-labels. These labels guide the training through a feedback loop.}
    \label{fig2}
\vspace{-10pt}
\end{figure*}
\subsection{Framework Overview}

We propose a \textbf{Feature-Induced Structural Optimal Transport (FIS-OT)} framework for unsupervised action segmentation, designed to resolve the tension between the lack of discriminative features and the rigidity of temporal priors in conventional self-training methods. Our architecture consists of two Collaboratively optimized components: a \textbf{Feature Enhanced Generator} module and a \textbf{Feature-Induced OT decoder}.

The FEG module functions as a local feature encoder, employing a triplet loss strategy \cite{19tsa} to mine intrinsic temporal and semantic similarities between frames, thereby providing low-level supervision independent of pseudo-labels. The OT decoder, operating globally, dynamically constructs a structural cost matrix using the features learned by FEG and solves a Gromov-Wasserstein (GW) problem to generate globally consistent pseudo-labels.

These components are trained via a dual-level cyclic optimization mechanism: FEG enhances the local structure of the feature space to provide accurate structural guidance for the OT, while the global temporal constraints imposed by the OT solution, in turn, guide the evolution of the feature space. This effectively breaks the confirmation bias often observed in self-training loops.
\vspace{-5pt}
\subsection{Temporal-Semantic Representation Learning}

\vspace{0.1cm}
\noindent\textbf{Problem formulation.} Given a sequence of frame-level features $X = \{x_1, \dots, x_N\} \in \mathbb{R}^{N \times D_{in}}$ for an untrimmed video, we first map them to a normalized latent space $Z = \{z_1, \dots, z_N\} \in \mathbb{R}^{N \times D}$ via a shallow encoder $f_\theta$, such that $\|z_i\|_2 = 1$. To prevent feature collapse in the unsupervised setting and capture local action continuity, we introduce the FEG module as an intrinsic regularization.

\vspace{0.1cm}
\noindent\textbf{Hybrid Similarity Measure.} We define a target similarity distribution $P_{target}(j|i)$ between frame $i$ and frame $j$ as a weighted fusion of temporal proximity and semantic similarity:

\begin{equation}
    S_{ij} = \alpha \cdot \underbrace{\exp\left(-\frac{|i-j|}{\sigma_t}\right)}_{\text{Temporal}} + (1-\alpha) \cdot \underbrace{\exp\left(-\frac{1 - z_i^\top z_j}{\sigma_s}\right)}_{\text{Semantic}}.
\end{equation}

where $\sigma_t$ and $\sigma_s$ denote the Gaussian kernel bandwidths for temporal and semantic terms, respectively.

\vspace{0.1cm}
\noindent\textbf{Triplet Selection \& Loss.} Unlike traditional supervision based on noisy pseudo-labels, FEG optimizes directly in the feature space. For each anchor frame $i$, we sample a set of positive samples $\mathcal{P}_i$ (top-K highest similarity) and a set of negative samples $\mathcal{N}_i$ (semi-hard samples). The FEG objective is formulated as a KL-divergence-based triplet loss:

\begin{equation}
    \mathcal{L}_{FEG} = \frac{1}{N} \sum_{i=1}^N \max\left(0, \text{KL}(S_i || S_{\mathcal{P}_i}) - \text{KL}(S_i || S_{\mathcal{N}_i})\right).
\end{equation}

This loss forces the encoder to pull semantically consistent frames closer while pushing dissimilar ones apart within a local neighborhood, establishing a robust feature foundation for the subsequent global segmentation.
\subsection{Feature-Induced Residual Structural OT}

This component represents the core contribution of our approach. We model action segmentation as a sequence alignment problem, seeking an optimal transport plan $T \in \mathbb{R}^{N \times K}$ that assigns $N$ video frames to $K$ action prototypes $A \in \mathbb{R}^{K \times D}$. We adopt the \textbf{Fused Gromov-Wasserstein (FGW)} distance as our objective:

\begin{equation}
\begin{split}
    \min_{T \in \Pi} \quad & (1-\beta) \underbrace{\langle C^k, T \rangle}_{\text{Matching Cost}} + \beta \underbrace{\sum_{i,j,k,l} L(C^v_{ij}, C^a_{kl}) T_{ik} T_{jl}}_{\text{Structural Cost}} \\
    & - \epsilon H(T) +\rho \text{D}_{\text{KL}}(T^\top \mathbf{1}_N || \mathbf{q}).
\end{split}
\end{equation}

\vspace{0.1cm}
\noindent\textbf{Matching Cost ($C^k$).} The Matching Cost measures the cosine distance between frame embeddings and action prototypes: $C^k_{ik} = 1 - z_i^\top a_k$.

\vspace{0.1cm}
\noindent\textbf{Residual Structural Prior ($C^v$).} Existing OT-based methods, such as ASOT, typically employ a fixed band matrix to enforce temporal consistency, which ignores non-linear variations within actions. We propose a \textbf{residual} dynamic structure that integrates a rigid temporal skeleton with dynamic feature affinity:

\begin{equation}
    C^v_{ij} = \underbrace{\mathbb{I}(|i-j| < Nr)}_{\text{Temporal Mask}} \odot \left( \mathbf{1} + \lambda \cdot \underbrace{\exp\left(-\frac{1 - z_i^\top z_j}{h}\right)}_{\text{Feature Affinity}} \right) \cdot \frac{1}{r}
\end{equation}

This formulation balances the stability and plasticity aspects that described as follows.

\subsubsection{Stability} 
The indicator function $\mathbb{I}(\cdot)$ and the base term $\mathbf{1}$ form a hard temporal window, preventing structural collapse during early training stages when features are ambiguous.

\subsubsection{Plasticity}
The feature affinity term allows the OT solver to perceive the internal structure of actions (e.g., repetitive sub-actions). When two frames fall within the temporal window and share high feature similarity, their connection strength is enhanced, promoting their assignment to the same action cluster.

Notably, the feature affinity utilizes the same Gaussian kernel form as the FEG loss, ensuring metric consistency between local optimization and global constraints.

\vspace{0.1cm}
\noindent\textbf{Action Mutual Exclusivity ($C^a$).} To prevent mode collapse, we enforce a mutual exclusivity structure among action prototypes, defined as $C^a_{kl} = 1 - \delta_{kl}$. This imposes a heavy structural penalty on assigning similar frames to different action clusters.

The problem is solved efficiently using the log-domain stabilized Sinkhorn algorithm to obtain the soft pseudo-labels $Q = T^*$.
\vspace{-5pt}
\subsection{Curriculum Optimization Strategy}

Since the feature-induced structure relies on high-quality representations, directly solving the OT with randomly initialized features may lead to a cold-start problem. We design a two-stage curriculum learning strategy to mitigate this:

\subsubsection{Structural Warm-up Phase} 
In the early stages ($t < T_{warm}$), we set the feature guidance coefficient $\gamma = 0$. Under this setting, the model operates similarly to ASOT, utilizing strong, fixed temporal priors to learn a coarse action skeleton.

\subsubsection{Refinement Phase} 
After warm-up ($t \ge T_{warm}$), we enable feature guidance ($\gamma > 0$). The fine-grained features learned by FEG begin to refine the OT structural matrix ($C^v$), allowing the segmentation boundaries to better align with the semantic manifold of the video content.

The final training objective combines the cross-entropy loss between the network prediction $P$ and the OT pseudo-labels $Q$, along with the FEG regularization:
\begin{equation}
    \mathcal{L}_{total} = \mathcal{L}_{CE}(P, Q) + \gamma(t) \cdot \mathcal{L}_{FEG}.
\end{equation}
\section{Experiments}
\begin{table*}[t]
\centering
\caption{State-of-the-art comparison on three unsupervised action segmentation benchmarks (Activity-Level). All numbers are \textbf{Full} metrics (\%). \textbf{Bold} indicates the best performance, and \underline{underline} indicates the second best. $^\dagger$: Our reproduction.}
\label{tab:1}
\resizebox{0.87\linewidth}{!}{%
\begin{tabular}{l|ccc|ccc|ccc|ccc}
\hline
 & \multicolumn{3}{c|}{Breakfast} & \multicolumn{3}{c|}{50 Salads (Mid)} & \multicolumn{3}{c|}{50 Salads (Eval)} & \multicolumn{3}{c}{Desktop Assembly} \\ 
\cline{2-13} 
Method & MoF & F1 & mIoU & MoF & F1 & mIoU & MoF & F1 & mIoU & MoF & F1 & mIoU \\
\hline
CTE \cite{8cte} & 41.8 & 26.4 & -- & 30.2 & -- & -- & 35.5 & -- & -- & 47.6 & 44.9 & -- \\
VTE \cite{17vte} & 48.1 & -- & -- & 24.2 & -- & -- & 30.6 & -- & -- & -- & -- & -- \\
UDE \cite{18ude} & 47.4 & 31.9 & -- & -- & -- & -- & 42.2 & 34.4 & -- & -- & -- & -- \\
ASAL \cite{9asal} & 52.5 & 37.9 & -- & 34.4 & -- & -- & 39.2 & -- & -- & -- & -- & -- \\
TOT \cite{2tot} & 47.5 & 31.0 & -- & 31.8 & -- & -- & 47.4 & 42.8 & -- & 56.3 & 51.7 & -- \\
TOT+ \cite{2tot} & 39.0 & 30.3 & -- & 34.3 & -- & -- & 44.5 & 48.2 & -- & 58.1 & 53.4 & -- \\
UFSA \cite{5ufsa} & 52.1 & 38.0 & -- & 36.7 & 30.4 & -- & \underline{55.8} & 50.3 & -- & \textbf{65.4} & \underline{63.0} & -- \\
HVQ \cite{4hvq} & \underline{54.4} & \textbf{39.7} & -- & -- & -- & -- & -- & -- & -- & -- & -- & -- \\
ASOT$^\dagger$\cite{6asot} & 51.0 & 34.6 & \underline{13.2} & \underline{41.9} & \underline{32.3} & \underline{20.3} & \textbf{59.6} & \textbf{52.9} & \underline{29.3} & 56.3 & 58.2 & \underline{35.5} \\
\hline
\textbf{Ours} & \textbf{57.9} & \underline{38.8} & \textbf{17.3} & \textbf{44.9} & \textbf{39.0} & \textbf{24.8} & 49.2 & \underline{52.3} & \textbf{34.5} & \underline{63.1} & \textbf{71.7} & \textbf{45.4} \\
\hline
\end{tabular}
}
\vspace{-10pt}
\end{table*}
\begin{table*}[t]
\centering
\caption{Comparison of video-level results on three benchmarks. \textbf{Bold} indicates the best performance, and \underline{underline} indicates the second best. $^\dagger$: Our reproduction.}
\label{tab:2}
\resizebox{0.87\linewidth}{!}{%
\begin{tabular}{l|ccc|ccc|ccc|ccc}
\hline
 & \multicolumn{3}{c|}{Breakfast} & \multicolumn{3}{c|}{50 Salads (Mid)} & \multicolumn{3}{c|}{50 Salads (Eval)} & \multicolumn{3}{c}{Desktop Assembly} \\ 
\cline{2-13}
Method & MoF & F1 & mIoU & MoF & F1 & mIoU & MoF & F1 & mIoU & MoF & F1 & mIoU \\
\hline
TWF \cite{15twf} & 62.7 & 49.8 & \textbf{42.3} & 66.8 & \textbf{56.4} & \textbf{48.7} & \textbf{71.7} & -- & -- & \textbf{73.3} & 67.7 & \textbf{57.7} \\
ABD \cite{16abd} & \underline{64.0} & \underline{52.3} & -- & \textbf{71.8} & -- & -- & -- & -- & -- & -- & -- & -- \\
ASOT$^\dagger$\cite{6asot} & 59.6 & 46.3 & 26.1 & 61.2 & 45.9 & 29.9 & \underline{62.3} & \underline{56.5} & \underline{30.2} & 68.5 & \underline{71.5} & 47.5 \\
\hline
\textbf{Ours} & \textbf{65.4} & \textbf{52.4} & \underline{32.4} & \underline{67.7} & \underline{55.4} & \underline{36.7} & 58.8 & \textbf{56.6} & \textbf{31.6} & \underline{71.8} & \textbf{78.4} & \underline{54.0} \\
\hline
\end{tabular}
}
\vspace{-10pt}
\end{table*}

\subsection{Experimental Setting}
To validate the effectiveness of our approach and ensure a fair comparison with prior state-of-the-art (SOTA) methods, we evaluate our framework on three widely used action segmentation benchmarks: \textbf{Breakfast}, \textbf{50Salads}, and \textbf{Desktop Assembly (DA)}. See Appendix for datasets details and additional studies. 
We follow the standard evaluation protocol for unsupervised action segmentation\cite{1cad,2tot,3otas,4hvq,5ufsa,6asot}, we apply Hungarian Matching at either the video level or activity level to establish a mapping between predicted clusters and ground truth labels. We report three core metrics:
\begin{itemize}
    \item \textbf{Mean over Frames (MoF):} This metric measures the percentage of correctly predicted frames. It is a standard evaluation metric. However, it is biased towards long action classes. It tends to overlook errors in short segments due to class imbalance.
    \item \textbf{F1 Score (F1):} We report the segmental F1 score. It is the harmonic mean of precision and recall. This metric evaluates the quality of segmentation boundaries. It effectively penalizes over-segmentation errors and assesses the detection of short actions.
    \item \textbf{Mean Intersection over Union (mIoU):} This metric calculates the overlap between predictions and ground truth. It averages the score across all action categories. By treating each class equally, it provides a robust evaluation for datasets with unbalanced class distributions.
\end{itemize}
\vspace{-5pt}
\subsection{Implementation Details}
The encoder employs a two-layer MLP, with output dimension varying according to the dataset. We use the Adam optimizer for training. The learning rate is set to $10^{-3}$ and weight decay to $10^{-4}$. For the Optimal Transport solver, we adopt the Mirror Descent algorithm to optimize the Gromov-Wasserstein objective. At each iteration, we apply a log-domain stabilized Sinkhorn projection. Furthermore, we integrate a Feature-Induced Hybrid Prior into the solver to unify metric learning and structural decoding. All experiments run on a single NVIDIA RTX A6000 GPU.
\vspace{-5pt}
\subsection{Comparison to State-of-the-Art}
We provide a comprehensive comparison of our proposed method against existing mainstream unsupervised frameworks, including classical clustering-based methods (CTE\cite{8cte}, ASAL\cite{9asal}), temporal optimal transport-based methods (TOT\cite{2tot}, UFSA\cite{5ufsa}), and recent SOTA approaches such as ASOT\cite{6asot} and HVQ\cite{4hvq}. Quantitative results in Table~\ref{tab:1} and Table~\ref{tab:2} present the detailed comparison results across the three datasets.

\vspace{0.15cm}
\noindent\textbf{Overall Performance.} As shown in Tables~\ref{tab:1} and~\ref{tab:2}, FIS-OT consistently achieves superior or competitive results. Notably, on the \textbf{Desktop Assembly} dataset, we outperform the second-best method by a large margin in F1 score (\textbf{71.7\%} vs. 63.0\%). This confirms that our feature-induced structure captures action transitions effectively without relying on rigid ordering assumptions.
In terms of Video-Level metrics, our method ranks second on the 50Salads (Mid) dataset. This is mainly due to the severe occlusions and local noise in the frames. Optimal Transport relies on global matching, which makes it sensitive to noise within individual videos. Although our method improves upon ASOT, it is still based on the Optimal Transport framework, limiting its performance in this specific scenario.

\vspace{0.15cm}
\noindent\textbf{Short Action Detection.} Our method excels in detecting short actions, particularly on the \textbf{50Salads (Mid)} split. We achieve a \textbf{+6.7\%} gain in F1 score over the ASOT baseline (39.0\% vs. 32.3\%). ASOT often misses short transitional segments due to over-smoothing. In contrast, our FEG module preserves these fine-grained details by mining intrinsic local semantic cues.

\vspace{0.15cm}
\noindent\textbf{Mitigating Length Bias.} Conventional OT methods often struggle with length bias and low boundary precision. Our approach significantly improves mIoU, showing gains of \textbf{+4.1\%} on Breakfast and \textbf{+5.2\%} on 50Salads (Eval) compared to ASOT. This indicates that our dynamic constraints generate segments that align better with the true action distribution.


\vspace{-5pt}
\subsection{Ablation Study}
\begin{table}[t]
\centering
\caption{Ablation study of key components on the \textbf{Desktop Assembly} dataset. We report results using both Video-Level and Activity-Level metrics (\%). ``w/o'' denotes removing a specific module. Best results are in \textbf{bold}.}
\label{tab:ablation_da}
\resizebox{\linewidth}{!}{%
\begin{tabular}{l|ccc|ccc}
\hline
 & \multicolumn{3}{c|}{Video-Level} & \multicolumn{3}{c}{Activity-Level} \\ 
\cline{2-7}
Method Variant & MoF & F1 & mIoU & MoF & F1 & mIoU \\
\hline
w/o Feature Residual & 68.8 & 74.1 & 49.9 & 60.5 & 65.4 & 41.2 \\
w/o Warm up & 70.7 & 75.5 & 52.4 & 58.5 & 60.6 & 38.9 \\
w/o FEG module & 68.5 & 71.5 & 47.5 & 56.3 & 58.2 & 35.5 \\
\hline
\textbf{Full Model (FIS-OT)} & \textbf{71.8} & \textbf{78.4} & \textbf{54.0} & \textbf{63.1} & \textbf{71.7} & \textbf{45.4} \\
\hline
\end{tabular}
}
\vspace{-15pt}
\end{table}
To understand the contribution of each component, we performed a systematic ablation study on the Desktop Assembly dataset (see Table~\ref{tab:ablation_da}). We report both Video-Level and Activity-Level metrics to evaluate local consistency and global semantic understanding.

\paragraph{Effect of Feature Enhanced Generator (FEG)}
Removing the FEG module leads to the largest performance drop. Specifically, the Activity-Level F1 score falls by \textbf{13.5\%} (71.7\% vs. 58.2\%). This shows that relying only on global OT alignment is not enough. Without the FEG, the model struggles to handle noisy video features. The FEG provides essential semantic cues that are necessary for high-quality segmentation.

\paragraph{Impact of Feature-Induced Structure}
To test our dynamic structure, we replaced it with a fixed temporal mask (w/o Feature Residual). This results in a \textbf{6.3\%} decrease in Activity-Level F1. This confirms that fixed temporal priors are too rigid. They cannot adapt to actions with different speeds. Our feature-induced approach gives the model flexibility. It allows the solver to adjust segmentation boundaries based on the actual video content.

\paragraph{Necessity of Warm-up Strategy}
We also analyzed the impact of our curriculum training. Skipping the warm-up phase (w/o Warm up) causes a sharp decline in performance. The Activity-Level F1 drops to 60.6\%. This suggests that a stable start is crucial. We must establish a coarse structural skeleton using fixed priors before we can effectively refine it with feature cues.

\subsection{Qualitative Results And Analysis}
\begin{figure}[htbp]
    \centering
    \includegraphics[width=.98\linewidth]{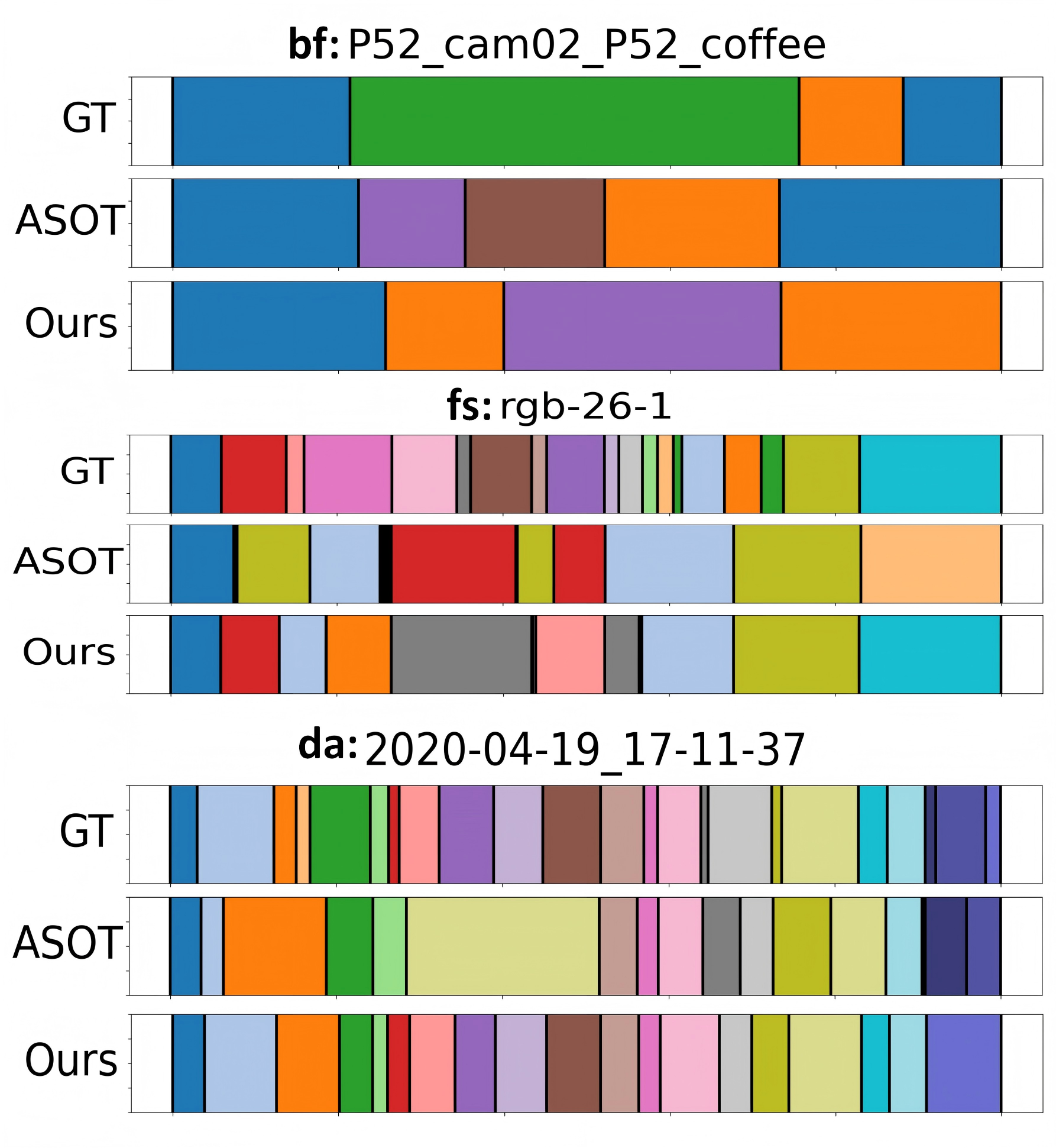}
    \vspace{-0.2cm}
    \caption{Qualitative results. We display the Ground Truth (GT), the predictions of FIS-OT (Ours), and the baseline ASOT. We show examples from different datasets and activities for comparison. The color-coded bars represent different action categories.}
    \label{qualitative}
\vspace{-5pt}
\end{figure}
Figure~\ref{qualitative} visualizes the results on the Breakfast, 50Salads, and Desktop Assembly datasets. On \textbf{Breakfast}, our model reduces the segmentation noise found in ASOT and produces more consistent predictions. In the \textbf{50Salads} example, ASOT tends to merge continuous steps into large blocks. Our method does a better job than ASOT by correctly identifying the action at the beginning and end of the video. The success of our approach is most evident on \textbf{Desktop Assembly}. As shown in the middle section of the timeline, ASOT fails to recognize the sequence of short actions and outputs a single long segment. In contrast, our method successfully detects these fine-grained details. It recovers the correct flow of short assembly steps, aligning closely with the ground truth.

\section{Conclusion}

We proposed FIS-OT, a feature-induced optimal transport framework for unsupervised action segmentation. By coupling local representation learning with global structural alignment in a cyclic optimization process, FIS-OT overcomes the rigid priors and confirmation bias present in prior OT-based methods. Extensive experiments on Breakfast, 50Salads, and Desktop Assembly demonstrate consistent competitive and state-of-the-art performance, with clear improvements in boundary accuracy and short-action segmentation. These results show that integrating feature-driven structure into optimal transport provides a more adaptive and reliable solution for unsupervised video segmentation.
\vspace{-5pt}
\section*{Acknowledgment}
This work was supported in part by the Natural Science Foundation of Jiangsu Province Basic Research Program under Grant BK20250489; in part by the the NSF of China under Grant 62502492; and in part by the Open Project Program of Shanghai Key Laboratory of Data Science (No.2025090600006).
\bibliographystyle{IEEEtran}
\bibliography{mybib}
\end{document}